\documentclass{article}

\PassOptionsToPackage{numbers, sort}{natbib}
\usepackage[final,nonatbib]{neurips_2024}

\usepackage[utf8]{inputenc} % allow utf-8 input
\usepackage[T1]{fontenc}    % use 8-bit T1 fonts
\usepackage{hyperref}       % hyperlinks
\usepackage{url}            % simple URL typesetting
\usepackage{booktabs}       % professional-quality tables
\usepackage{amsfonts}       % blackboard math symbols
\usepackage{nicefrac}       % compact symbols for 1/2, etc.
\usepackage{microtype}      % microtypography
\usepackage{xcolor}         % colors
\usepackage{amsmath}
\usepackage{graphicx}

\title{DRACO: A Denoising-Reconstruction Autoencoder for Cryo-EM}

\newcommand*\samethanks[1][\value{footnote}]{\footnotemark[#1]}
\author{
Yingjun Shen$^{1,2\thanks{The authors contributed equally to this work.}}$ ~~
Haizhao Dai$^{1,2\samethanks[1]}$ ~~
Qihe Chen$^{1,2}$ ~~
Yan Zeng$^{1,2}$ \And
Jiakai Zhang$^{1,2}$ ~~
Yuan Pei$^{1,3}$ ~~
Jingyi Yu$^{1}$ \\
$^1$School of Information Science and Technology, ShanghaiTech University. \\
$^2$Cellverse Co, Ltd. \\
$^3$iHuman Institute, ShanghaiTech University. \\
{\tt\small \{shenyj2022,daihzh2023\}@shanghaitech.edu.cn} \\
{\tt\small \{chenqh2024,zengyan2024,\newline
zhangjk,peiyuan,yujingyi\}@shanghaitech.edu.cn}
}

\usepackage{xspace}

\newcommand{\ours}{DRACO\xspace}

\begin{document}

\maketitle
\bibliographystyle{unsrt}
\begin{abstract}
Foundation models in computer vision have demonstrated exceptional performance in zero-shot and few-shot tasks by extracting multi-purpose features from large-scale datasets through self-supervised pre-training methods.
However, these models often overlook the severe corruption in cryogenic electron microscopy (cryo-EM) images by high-level noises. We introduce \ours, a Denoising-Reconstruction Autoencoder for CryO-EM, inspired by the Noise2Noise (N2N) approach. By processing cryo-EM movies into odd and even images and treating them as independent noisy observations, we apply a denoising-reconstruction hybrid training scheme. We mask both images to create denoising and reconstruction tasks.
For \ours's pre-training, the quality of the dataset is essential, we hence build a high-quality, diverse dataset from an uncurated public database, including over 270,000 movies or micrographs. After pre-training, \ours naturally serves as a generalizable cryo-EM image denoiser and a foundation model for various cryo-EM downstream tasks. \ours demonstrates the best performance in denoising, micrograph curation, and particle picking tasks compared to state-of-the-art baselines. 
\end{abstract}

\section{Introduction}

Foundation models in computer vision have demonstrated remarkable capabilities in zero-shot and few-shot tasks.
These models learn to extract multi-purpose visual features from large-scale, diverse datasets through text-guided \cite{clip, flip, coca} or self-supervised \cite{ibot,dinov2} pre-training methods such as masked image modeling (MIM) \cite{mae}.
The features can then be applied to various downstream tasks.
For instance, DINOv2 \cite{dinov2} is trained on a large-scale curated dataset and shows significant performance improvements in classification, retrieval, segmentation, etc.
The success of vision foundation models has stimulated advances across various scientific disciplines.
Due to the diverse modalities of scientific imaging, training domain-specific foundation models \cite{uni, conch, unifmir, retfound} is essential to meet specific demands.
For example, the UNI \cite{uni} foundation model for tissue imaging is pre-trained on 100 million images for 34 representative clinical downstream tasks.
% Similarly, the foundation model of fluorescence microscopy imaging (UniFMIR) has been trained on a multimodal dataset derived from structured illumination microscopy (SIM) and direct stochastic optical reconstruction microscopy (dSTORM) for generalized fluorescence image restoration.

\begin{figure}[t]
    \centering
    \includegraphics[width=\linewidth]{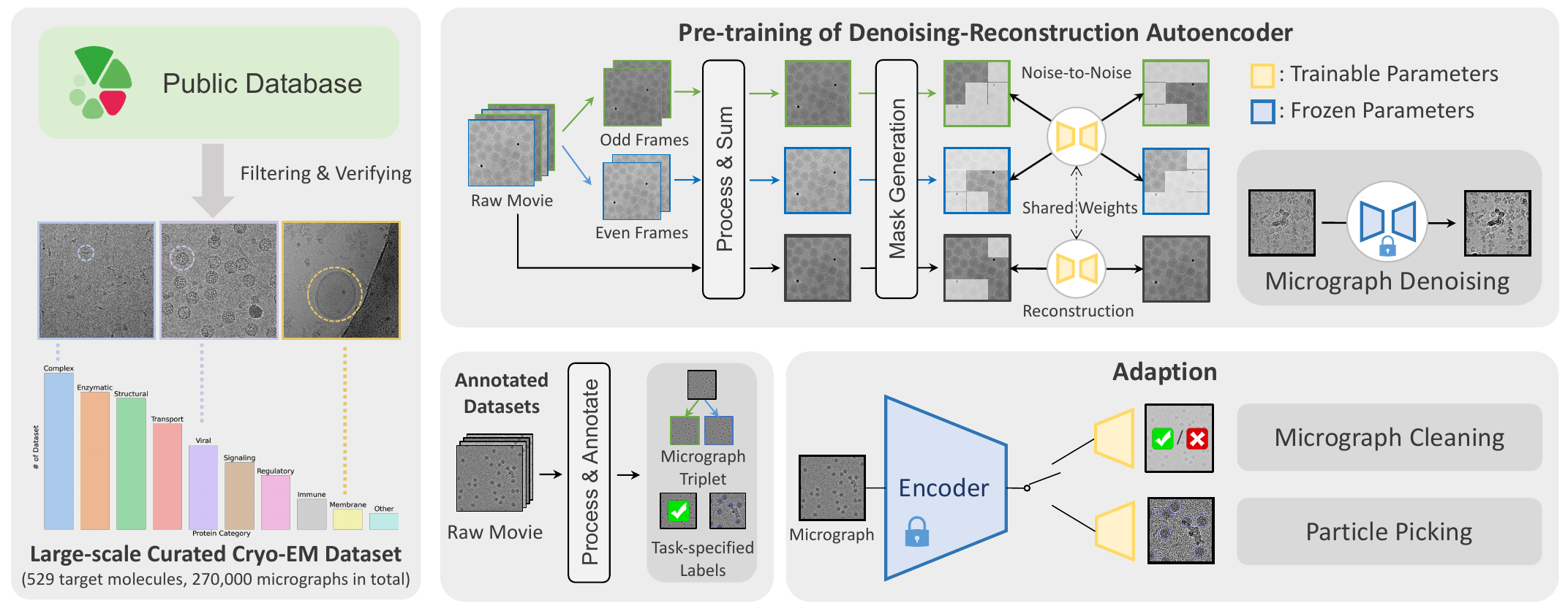}
    \caption{\textbf{Overview of \ours.} For pre-training, we construct a large-scale curated dataset containing 529 types of protein data with over 270,000 cryo-EM movies or micrographs. Based on this, we present \ours, a denoising-reconstruction autoencoder for cryo-EM. A pre-trained \ours naturally serves as a generalizable cryo-EM image denoiser and a foundation for various downstream model adaptions such as micrograph curation and particle picking. }
    \label{fig:overview}
    \vspace{-15pt}
\end{figure}

In structural biology, cryogenic electron microscopy (cryo-EM) stands as a pivotal bio-imaging technique \cite{ResRev}.
Unlike optical imaging methods, cryo-EM possesses several distinctive characteristics:
first, cryo-EM utilizes high-energy electron beams as its illumination source \cite{cryo-em-formation} and direct detector device (DDD) captures a continuous multi-frame sequence, often called a \textit{movie} \cite{motioncor2}.
To mitigate specimen damage during exposure, the electron dose per frame is restricted, leading to \emph{extremely low signal-to-noise ratios (SNR)} in the captured images.
Second, cryo-EM employs motion correction to counteract blurring induced by specimen drift during exposure to ultimately obtain a sharper single micrograph \cite{motioncor2}.
Last, the acquired images comprise hundreds of thousands of target protein particles with diverse poses. 
%To resolve 3D structures, researchers utilize methods grounded in multi-view reconstruction \cite{cryodrgn,cryo-fire}.
To resolve 3D structures, researchers utilize a pipeline composed of multiple downstream tasks including micrograph curation, micrograph denoising, particle picking, pose estimation, and ultimately high-resolution reconstruction.

% 现有的自监督图像预训练方法，比如MAE，通过补全被遮盖图像中的丢失信息，提取干净的自然图像中的通用特征。然而冷冻电镜的图像被high-level noise破坏，这种方法补全的信号往往严重的丢失高频信息，暗示高频信息并没有被成功捕捉。
% 实际上，冷冻电镜的

% Aligned with denoising autoencoders~\cite{dae}, which value ``robustness to partial destruction of the input" as a good representation criterion, 

In line with denoising autoencoders \cite{dae}, which consider ``robustness to partial destruction of the input'' a good representation criterion, existing self-supervised learning methods, such as masked autoencoders (MAE) \cite{mae}, have been successful in learning expressive representations by reconstructing the missing patches from a partially masked image.
However, in cryo-EM, these methods overlook the severe corruption caused by pixel-level random noise, leading to degraded performance.
To be more robust to noises, DMAE \cite{dmae} reconstructs the clean image from the masked one that is further corrupted by synthetic Gaussian noise.
Nevertheless, cryo-EM clean reference images are impossible to obtain due to the fragile biological specimens, posing a significant challenge.
% In line with denoising auto-encoders~\cite{dae}, a good criterion for a representation is its ``robustness to partial destruction of the input''. Driven by this, existing self-supervised learning methods, such as masked auto-encoders (MAE) ~\cite{mae}, attempt to learn an expressive representation by randomly masking patches from input images and reconstructing the missing ones. However, in cryo-EM, these methods overlook the severe corruption of images caused by random noise at the pixel level, leading to their degraded performance. DMAE~\cite{dmae} further accounts for the noisy image by reconstructing the clean image from the masked one corrupted by synthetic Gaussian noise. Nonetheless, cryo-EM lacks clean reference images, posing a significant challenge.
% TODO:
% XXX: no good, need to be rephrased.
% Due to the signals in cryo-EM images being severely disrupted by noise, existing methods like MAE \cite{mae} are ineffective in recovering high-frequency signals, leading to reduced downstream task performance. Denoising MAE \cite{dmae} and DDeP \cite{ddep} add synthetic Gaussian noise to clean images to train a model that reconstructs the clean image from the corrupted one. However, cryo-EM lacks clean reference images. Noise2Noise (N2N) \cite{noise2noise} denoises images by learning the mappings between the paired noisy images with zero-mean noise, but XXX. 

In this paper, we present \ours, a Denoising-Reconstruction Autoencoder for CryO-EM, as shown in Figure \ref{fig:overview}. Inspired by Noise2Noise (N2N) \cite{noise2noise}, which learns to denoise images using only paired noisy images, we divide the original movie into two sub-movies based on odd and even frame numbers, processing them into odd and even images. We treat them as two independent, noisy observations of the underlying true signal, thus the idea of N2N can apply. During training, we partially mask both images, creating masked regions and unmasked regions corresponding to denoising and reconstruction tasks: in the unmasked region, the odd noisy patch learns to recover the even noisy patch, and vice versa. In the masked region, we introduce relatively low-noise images from the complete movie to supervise the reconstructed results. This denoising-reconstruction hybrid training scheme achieves the robust feature extraction of noisy cryo-EM images.

The quality of training samples is crucial for the general-purpose feature extraction of cryo-EM images.
Direct access to the public database leads to varying data quality, inconsistent data formats, or missing annotations.
Therefore, we construct a large-scale, high-quality, and diverse single-particle cryo-EM image dataset by curating and manually processing 529 sets of data from EMPIAR \cite{empiar}, obtaining over 270,000 cryo-EM movies or micrographs in total. % For each movie, we conduct motion correction \cite{motioncor2} on three distinct versions: the odd movies, the even movies, and the original movies, to obtain three corresponding micrographs. 
After pre-training, \ours naturally serves as a generalizable cryo-EM image denoiser and a foundation for various downstream model adaptions. We hence explore the performance of \ours on three downstream tasks: micrograph curation, denoising, and particle picking. Extensive experiments show that \ours outperforms the state-of-the-art baselines in all downstream tasks. We will release code, pre-trained/fine-tuned models, and the large-scale curated dataset.

% We extensively evaluate \ours's performance. In denoising tasks, compared to the baseline, we achieved XXX improvement in XXX performance. In micrograph curation tasks, compared to unsupervised small models without pre-training, our fine-tuned approach attained XXX performance improvement. Finally, in particle picking tasks, we conducted zero-shot performance validation on three public datasets. Compared to traditional methods, we achieved XXX\% performance improvement on XXX metrics. We will open-source all pre-training, downstream task code, model parameters, and datasets, hoping to XXX.

\section{Related Work}
Our work aims to extend the vision foundation model to the field of cryo-EM. We therefore only discuss the most relevant works in respective fields.
% 【讲视觉领域的模型，引用上下游模型成功的例子。生命科学领域中，它们用了什么模型，主要针对什么图像。为什么我们要特殊设计】
% A foundation model is any model that is trained on broad data (generally using self-supervision at scale) that can be adapted (e.g., fine-tuned) to a wide range of downstream tasks.
% \paragraph{Vision foundation model} 通过在大规模图像数据集上 \cite{jft300m,laion-5b} 使用自监督学习方法进行预训练，旨在提取出能广泛应用于不同任务的通用视觉信号，以快速适应各种下游视觉任务。特别是基于 Vision Transformer \cite{vit} (ViT) 架构的自监督学习策略，这些策略致力于构建自然图像上的前置任务，如对比学习 \cite{clip,dinov2} 通过学习图像间或模态间的相似性，以及masked image modeling \cite{beitv2,ibot} 通过恢复被遮盖图像的相关特征来学习。在这些方法中，masked auto-encoder (MAE) \cite{mae} 特别地，通过仅重建low-level的图像像素，实现了对图像high-level语义的理解。其衍生模型在多个下游任务中展示了显著效果，包括图像分类、目标检测 \cite{vitdet} 和图像分割 \cite{sam}，证明了其强大的泛化能力。

\noindent\textbf{Vision foundation models in computer vision.}
Vision foundation models are pre-trained on large-scale image datasets \cite{jft300m, laion5b} using self-supervised learning methods \cite{ssl}, aimed at extracting general visual signals rapidly adaptable to various downstream visual tasks \cite{vitdet, sam,dust3r,foundation-model}.
Techniques for pre-training vision foundation models, such as contrastive learning \cite{simclr,infonce, clip,beitv3} and self-distillation \cite{dino,dinov2}, focus on aligning features across different models or modalities, while another method, masked image modeling \cite{mae,beit,beitv2,ibot}, reconstructs features from masked images to capture high-level visual semantics.
However, these existing vision foundation models are not directly applicable to cryo-EM imaging.
In particular, their application to cryo-EM imaging is limited by the high noise levels in micrographs, which degrade signal capture.
Therefore, we propose a denoising-reconstruction pre-training framework that is robust to highly noisy cryo-EM micrographs, making it suitable for specific downstream tasks in cryo-EM.

% 视觉基础模型已广泛成功应用于生命科学的成像领域。具体而言，像 Swin Transformer 这样的多种视觉基础模型已被直接用于多种科学图像处理任务，包括视网膜图像 \cite{retfound}、荧光显微镜图像 \cite{unifmir}、组织病理学图像 \cite{uni,conch} 和放射学图像 \cite{medmd}。这些模型在相应领域的下游任务中，如疾病诊断、病灶检测和图像修复方面，已证明极为有效。然而，现有的视觉基础模型并不直接适用于cryo-EM图像，特别是因为cryo-EM显微图的高噪声水平极大地干扰了模型对信号的捕捉，从而限制了其性能。因此，我们的研究工作旨在提升这些模型的抗噪声能力，以便更好地适应cryo-EM的特定下游任务。

\noindent\textbf{Vision foundation models in life science.}
The remarkable success of foundation models has extended to various life science imaging domains, including applications in retinal \cite{retfound}, fluorescence microscopy \cite{unifmir}, histopathology \cite{uni, conch}, and radiology imaging \cite{medmd}.
These models have shown considerable effectiveness in tasks such as disease diagnosis, lesion detection, and image restoration within these fields.
In contrast to these domains, which benefit from extensive and well-curated datasets supporting model training, the cryo-EM field lacks such resources.
To fill this gap, we have developed a well-curated, large-scale dataset specifically designed to support the training of cryo-EM foundation models, ensuring that they can be effectively applied in this specialized field.

% 【简要介绍cryo-EM噪声，讲Cryo-EM的multiframe成像特点引出基于N2N的topaz denoise等工作，讲这些模型的局限，引出我们的基础模型将N2N和MAE结合】
% Cryo-EM 技术通过多次低剂量拍摄来减轻电子对样本的损伤，但这也导致了极低的信噪比和复杂的噪声模式。
% 传统的去噪方法通常采用泊松-高斯等模型进行噪声建模，并依靠滤波来去除噪声。
% 然而，这些方法过度简化了噪声模式，可能导致高频信号细节的丢失。
%learning based方法充分利用了多帧数据生成奇偶帧图像，基于N2N框架不需要干净图像只需要noisy图像对。但现有方法都只在小规模数据集上训练，缺乏泛化性，且CNN-based网络结构对图像特征的提取能力也不足。我们进一步将 N2N 框架与 MAE 相结合，提出了去噪-重构预训练框架，不仅能天然地成为一个泛化的去噪器，也能提取更robust的特征应用到下游任务中。

% 最近，深度学习方法为 cryo-EM 图像去噪提供了新的思路，尤其是 Noise2Noise（N2N）和Noise2Void 等技术，它们无需依赖干净图像即可进行去噪。
% 例如，Topaz Denoise和Cryo-CARE等工作通过生成奇数和偶数显微照片的方式，充分利用了多帧数据，有效抑制噪声。
% 我们进一步将 N2N 框架与 MAE 相结合，提出了去噪-重构预训练框架，显著提高了模型在高噪声cryo-EM图像中的特征提取能力，超越了一般去噪任务的能力。

\noindent\textbf{Cryo-EM image denoising.}
% The Cryo-EM technique mitigates electron damage to the specimen by multiple low-dose exposures \cite{}, but this also results in extremely low signal-to-noise ratios and complex noise patterns.
To tackle the issues of low SNR and complex noise patterns in cryo-EM images, traditional denoising techniques often employ noise models like the Poisson-Gaussian model \cite{cryo-em-formation,tem} and rely on filtering methods \cite{wavelet-filter,bilateral-filter,non-local-means-filter} to denoise. 
However, these methods oversimplify the noise patterns, which can lead to the loss of high-frequency signal details.
Recently, NT2C \cite{nt2c} uses Generative Adversarial Network to learn the noise patterns for denoising, but it requires simulated datasets as the clean references.
Another series of learning-based methods \cite{topaz-denoise,cryocare} make full use of multi-frame data by generating odd and even images for denoising based on Noise2Noise (N2N) \cite{noise2noise,noise2void} framework.
These methods do not require clean images for denoising, but they still suffer from small-scale datasets and network architectures, which limit their generalizability.
In this paper, we propose \ours, pre-trained on a large-scale curated dataset, that can naturally serve as a generalizable denoiser for cryo-EM micrographs.

% Recently, deep learning methods \cite{cryo-gem,cryo-nerf} have provided new ideas for cryo-EM image denoising, especially techniques such as Noise2Noise (N2N) \cite{noise2noise} and Noise2Void \cite{noise2void}, which can denoise without clean images.
% In cryo-EM, methods like Topaz Denoise \cite{topaz-denoise} and Cryo-CARE \cite{cryocare} make full use of multi-frame data by generating odd and even micrographs for effective noise suppression.
% We further combine the N2N framework with MAE to propose a denoising-reconstruction pre-training framework, which significantly improves the feature extraction ability of the model in highly noisy cryo-EM images, surpassing the capabilities of general denoising tasks.

% 【一个有效的基础模型可以benefit很多cryo-EM的下游任务。CTF估计……。Picking……。现有方法（CSPARC）是怎么做的。】
% 一个有效的基础模型可以 benefit 很多 cryo-EM 的下游任务, 包括 micrograph curation、图像去噪和颗粒挑选等。
% Micrograph curation 旨在确保只有高质量的图像用于后续分析，但现有方法仍然需要人工查看。（Miffi）
% 在图像去噪方面，虽然 Topaz Denoise 等工具取得了显著的去噪效果，但这些方法往往缺乏泛化能力。
% 颗粒挑选任务需要从 micrograph 中识别和提取具有代表性的颗粒，是 SPA 的一项重要任务。
% 模板匹配等传统方法严重依赖先验信息，需要大量一次性人工干预。
% 基于学习的模型，如 Topaz Picking、crYOLO 和 CryoTransformer，提供了更简化的流程，但由于数据规模有限，仍然缺少泛化性。
% \ours 在大规模 cryo-EM 图像数据集上预训练，能有效地适应这些任务，并展示出强大的泛化能力。
\noindent\textbf{Downstream tasks in single particle analysis.}
An effective foundational model can benefit downstream tasks in cryo-EM, including micrograph curation and particle picking.
Micrograph curation aims to ensure that only high-quality images are selected for further analysis, yet current methods rely heavily on manual inspection \cite{miffi,cryosparc}.
% In terms of denoising, although tools like Topaz Denoise \cite{topaz-denoise} have achieved significant denoising results, these methods fall short in terms of generalization, limiting their broader applicability.
Particle picking involves identifying and extracting representative particles from micrographs, which is a critical task in the cryo-EM single particle analysis (SPA) reconstruction pipeline.
Traditional methods \cite{relion,eman2, warp}, such as template matching \cite{findem,gautomatch} and difference of Gaussians (DoG) method \cite{dog}, heavily rely on prior information and require substantial \textit{ad hoc} post-processing.
Learning-based models \cite{deeppicker,deepem,cryomae}, such as Topaz-Picking \cite{topaz-picking}, crYOLO \cite{cryolo}, and CryoTransformer \cite{cryotransformer}, offer more streamlined processes but still face challenges in generalizability due to the limited data scale.
Our \ours, pre-trained on large-scale cryo-EM image datasets, can effectively adapt to these tasks and demonstrate strong generalization capabilities.

% As shown in Fig. \ref{fig:overview}, we aim to build a cryo-EM foundation model that can multi-purpose robust features for cryo-EM micrographs in high-level noise.
% To achieve this, we propose a novel self-supervised denoising-reconstruction pre-training framework that combines the Noise2Noise framework with a masked auto-encoder model (see Sec. \ref{sec:derec}).
% Regarding pre-training data, we selected, standardized, and curated a large-scale unlabeled image dataset from the public cryo-EM database EMPIAR (see Sec. \ref{sec:pretraining-setup}).
% We then pre-trained a model, called \ours, and applied it to a series of downstream tasks in the single particle analysis (SPA) pipeline through methods such as linear probing and fine-tuning, demonstrating its applicability and generalizability (see Sec. \ref{sec:experiments}).
% Notably, for each downstream task, we select a subset of high-quality micrographs and precisely annotate a small sample dataset using professional tools.
% This careful curation and annotation process ensures the reliability and accuracy of our model in real-world applications.

\section{Preliminary: Imaging Formation Model}\label{sec:preliminary}

Cryo-EM uses a Direct Detector Device (DDD) camera \cite{tem} for their notably higher detective quantum efficiency (DQE) compared to traditional cameras. This allows recording the micrograph as a multi-frame movie rather than a single integrated exposure.
In this setup, a movie is a series of continuous multi-frame images, denoted as $\mathcal{M} = \{\hat{I}_i\}_{i=1}^{M}$, where each frame $\hat{I}_i$ is an independent observation of the true signal $I$.

Ideally, the imaging process in cryo-EM involves two main steps:
1) projecting the 3D density volume of the region of interest $V(x, y, z): \mathbb{R}^3 \to \mathbb{R}$ along the $z$-axis via weak-phase object approximation \cite{wpoa}, and
2) modulating the projection image with the Point Spread Function (PSF) $g$ of the cryo-EM optical lens, expressed as:
\begin{equation}
    I = g * \int V(x, y, z)\,\mathrm{d}z.
\end{equation}
where $I$ is considered as the true signal.

However, in practice, the captured frames suffer from extremely low SNR due to the limited electron dosage and the high sensitivity of DDD. The main noise source is Poisson (shot) noise from the detector, denoted as $\text{Poisson}(I)$, arising from the inherent uncertainty of the electron measurement \cite{cryo-em-formation}. We assume that the additional noise types like heat, readout, and dark current noise are collectively modeled as a zero-mean Gaussian noise $\mathcal{G}$ with an unknown variance $\sigma^2$ \cite{relion}:
\begin{equation}
    \hat{I} = \text{Poisson}(I) + \mathcal{G}.
\end{equation}

As the number of observations increases, the average of these movie frames converges to the true signal:
\begin{equation}
    \mathbb{E}[\hat{I}] = I \approx \frac{1}{M}\sum_{i=1}^{M}\hat{I}_i.
\end{equation}

We define the image noise $\epsilon$ as the difference between the captured and clean images for each frame. Thus, the expectation (mean value) of noise distribution is:
\begin{equation}
    \mathbb{E}[\epsilon] = \mathbb{E}[\hat{I}_i - I] = 0.
\end{equation}
Thus, we derive that the cryo-EM image noise is zero-mean. This conclusion is also aligned with existing cryo-EM reconstruction methods \cite{relion, cryosparc}, which directly assume that the noise distribution is an additive zero-mean Gaussian yet can still achieve high-resolution reconstruction. We are well aware that this derivation is relatively trivial compared to an actual theoretical analysis \cite{cryo-em-formation}, but this gives us intuitive guidance to integrate the N2N \cite{noise2noise} idea: learning to denoise images from solely noisy image pairs, into our pre-training framework. 

\noindent\textbf{Movie to micrograph triplets.}
Given an $M$-frame movie $\mathcal{M}$, we divide it into odd frames $\mathcal{M}^{\text{o}} = \{I_{2i-1}\}_{i=1}^{\lceil M/2 \rceil}$ and even frames $\mathcal{M}^{\text{e}} = \{I_{2i}\}_{i=1}^{\lfloor M/2 \rfloor}$.
An off-the-shelf motion correction method \cite{motioncor2} is then applied to correct cross-frame drifts in $\mathcal{M}$, $\mathcal{M}^{\text{o}}$, and $\mathcal{M}^{\text{e}}$.
By summing up the frames within each subset, we generate three micrographs with the same shape: the original micrograph $\mathbf{\hat{I}}$, odd micrograph $\mathbf{\hat{I}}^{\text{o}}$, and even micrograph $\mathbf{\hat{I}}^{\text{e}}$,
As aforementioned, all these micrographs are expected to reflect the true signal $I$ but corrupted by noises.

\section{Denoising-reconstruction Autoencoder}\label{sec:derac}
\begin{figure}[t]
    \centering
    \includegraphics[width=\linewidth]{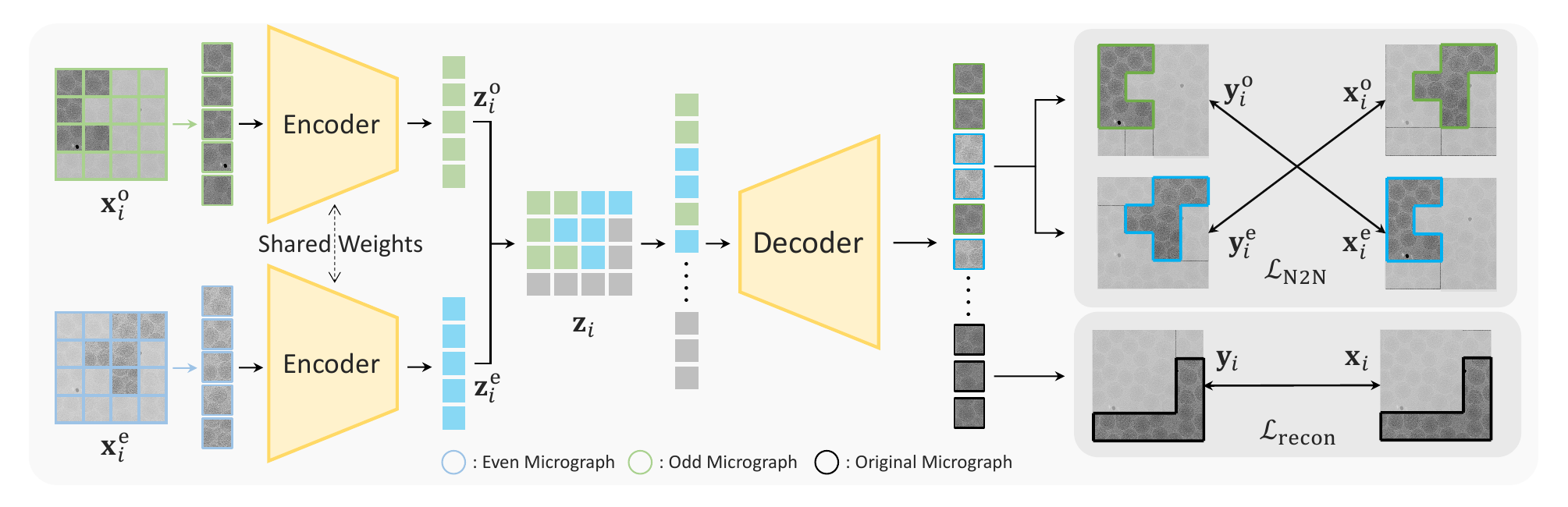}
    \caption{
        \textbf{The pipeline of \ours.}
        Given a pair of partially masked odd and even micrographs, the encoder takes odd-visible patches and even-visible patches as inputs. The unmasked latent patches are combined with masked latent patches together to generate the latent representation $\mathbf{z}_i$. Then the latent representation passes through the decoder to generate predicted patches. The N2N loss is applied to odd-visible predicted patches with corresponding even input patches, and vice versa. The reconstruction loss is applied to both invisible predicted patches with higher SNR input patches.
    }
    \label{fig:network}
\end{figure}

We introduce \ours, a denoising-reconstruction autoencoder for cryo-EM, as illustrated in Figure \ref{fig:network}. Different from existing masked imaging modeling methods, our model uses paired odd-even micrographs as inputs for the denoising target on visible patches. Further, we utilize the original micrographs as the additional supervision signal for reconstruction on masked patches.

\noindent\textbf{Masking.}
Following the standard scheme in Vision Transformer (ViT) \cite{vit}, each micrograph in a triplet, consisting of one original, one odd, and one even micrograph from the same movie, is divided into regular non-overlapping patches.
We create patch sets $\{\mathbf{x}^\text{o}_i\}_{i=1}^{N}$ for the odd, $\{\mathbf{x}^\text{e}_i\}_{i=1}^{N}$ for the even, and $\{\mathbf{x}_i\}_{i=1}^{N}$ for the original micrograph, where $N$ represents the number of patches.
For the odd and the even micrographs used as inputs in our model, we generate two sets of binary masks, $\{\mathbf{m}^{\text{o}}_i\}_{i=1}^{N}$ and $\{\mathbf{m}^{\text{e}}_i\}_{i=1}^{N}$, with a mask ratio $\gamma$.
Here, $\mathbf{m}_i = 1$ means the $i$-th patch is masked, and $0$ means unmasked. Additionally, we ensure that a patch can be 1) only visible in one of them, or 2) masked in both. This strategy ensures that each visible patch has no information sharing of its corresponding patch on the other micrograph. 
Notably, this requires $\gamma \ge 0.5$ for each input micrograph.

\noindent\textbf{Network architecture.}
For \ours's pre-training, we employ a ViT-based encoder-decoder architecture following the MAE framework \cite{mae}.
Positional embeddings are first added to the input patches, which are then masked to select only visible patches, denoted as $\{\mathbf{v}_i\}_{i=1}^{\gamma N}$.
The encoder, $G_{\text{enc}}$, is a ViT that transforms these visible patches from either odd or even micrographs into latent features.
To  align with the original unmasked size, zeros are padded to the latent features based on the positions indicated by the corresponding masks before the encoder outputs them:
\begin{equation}
    \{\textbf{z}^{\text{o}}_i\}_{i=1}^{N} = G_{\text{enc}}\left(\{\mathbf{v}^{\text{o}}_i\}_{i=1}^{\gamma N}, \{\mathbf{m}^{\text{o}}_i\}_{i=1}^{N}; \theta_{\text{enc}}\right),
    \quad
    \{\textbf{z}^{\text{e}}_i\}_{i=1}^{N} = G_{\text{enc}}\left(\{\mathbf{v}^{\text{e}}_i\}_{i=1}^{\gamma N}, \{\mathbf{m}^{\text{e}}_i\}_{i=1}^{N}; \theta_{\text{enc}}\right).
\end{equation}

Each of the latent features from odd and even micrographs retain part of the original micrograph's information.
To reconstruct the micrograph, we generate the latent representation $\{\mathbf{z}_i\}_{i=1}^{N}$ for further processing by the decoder:
\begin{equation}
    \mathbf{z}_i = (1-\mathbf{m}^{\text{o}}_i) \cdot \mathbf{z}^{\text{o}}_i + (1-\mathbf{m}^{\text{e}}_i) \cdot \mathbf{z}^{\text{e}}_i + \mathbf{m}^{\text{o}}_i \cdot \mathbf{m}^{\text{e}}_i \cdot \text{\texttt{[MASK]}},
\end{equation}
where the \texttt{[MASK]} token is a shared learnable embedding representing the masked patch to be predicted.
Finally, the decoder $G_{\text{dec}}$ takes the latent representation $\{\mathbf{z}_i\}_{i=1}^{N}$ with another positional embedding added as input and predicts all masked and visible patches to reconstruct the complete micrograph:
\begin{equation}
    \{\mathbf{y}_i\}_{i=1}^{N} = G_{\text{dec}}(\{\mathbf{z}_i\}_{i=1}^{N}; \theta_{\text{dec}}). 
\end{equation}

\noindent\textbf{Denoising target.}
Inspired by N2N and Topaz-Denoise \cite{topaz-denoise}, which predict denoised images only from paired noisy images, we introduce an image-denoising target on visible patches.
For any patch $\mathbf{x}_i$ visible in only the odd micrographs, the model predicts its counterpart in the even micrographs, and vice versa, as illustrated in Figure \ref{fig:network}.
As mentioned earlier, the expectation (mean) values of the odd and even micrographs are the true signal.
Therefore, following Topaz-Denoise, we employ a patch-wise L2 loss function for each visible patch, aiming for regressing the true signal:
\begin{equation}
    \mathcal{L}_{\text{N2N}}(\mathbf{x}_i^{\text{o}}, \mathbf{x}_i^{\text{e}}, \mathbf{y}_i) = 
    \begin{cases}
        \|\mathbf{x}_i^{\text{e}} - \mathbf{y}_i\|_2^2, & \text{if } \mathbf{m}^{\text{o}}_i = 0, \\
        \|\mathbf{x}_i^{\text{o}} - \mathbf{y}_i\|_2^2, & \text{if } \mathbf{m}^{\text{e}}_i = 0.
    \end{cases}
\end{equation}

\noindent\textbf{Reconstruction target.}
For masked patches, we let the decoder predict the pixel value of patches from the original micrograph with higher SNR compared to odd and even micrographs for better reconstruction quality. Thus, the reconstruction loss is:
\begin{equation}
    \mathcal{L}_{\text{recon}}(\mathbf{x}_i, \mathbf{y}_i) = \|\mathbf{x}_i - \mathbf{y}_i\|_2^2,\quad\text{if } \mathbf{m}_i^o \cdot \mathbf{m}_i^e = 1.
\end{equation}

\noindent\textbf{Training objective.}
During training, we combine the N2N loss with the reconstruction loss:
\begin{equation}
    \mathcal{L} = \mathcal{L}_{\text{N2N}} + \lambda\mathcal{L}_{\text{recon}}, 
\end{equation}
where $\lambda$ is a hyper-parameter set to $1.0$ in all our experiments.

% \section{Downstream Tasks}

\section{Experiments}\label{sec:experiments}

% We evaluate \ours in three downstream tasks
% 抗噪的预训练策略使得 \ours 可以非常容易地适应到 cryo-EM single particle analysis (SPA) 重建管线的关键下游任务中。
% 以下简要介绍这些任务及 \ours 的适应策略。
% % https://www.ncbi.nlm.nih.gov/pmc/articles/PMC9825465/pdf/gkac1062.pdf
% EMPIAR~\cite{Iudin2022EMPIARTE}(empiar.org) 是存储原始cryo-EM图像数据和vEM、XT实验的3D重建的公共档案库,现在包含超过2000个条目，总计超过2PB的数据。我们从生物学分析的角度定义数据集的质量，高质量的数据集应该能够产出高分辨的3D重建以进行后续分析，我们最终选择满足重建分辨率优于10\AA 的高质量图像数据集。
\noindent\textbf{Large-scale curated dataset for pre-training.}
The effectiveness and robustness of \ours depend heavily on the quality and scale of the cryo-EM pre-training dataset.
However, direct access to public databases like the Electron Microscopy Public Image Archive (EMPIAR) \cite{empiar} results in variations in data quality, inconsistent data formats, and inaccurate or even missing annotations.
To overcome these challenges, we have developed a data generation workflow.
First, we select datasets with reported resolutions better than 10 \AA, ensuring high-quality data acquisition.
Next, we collect the raw data, including metadata, movies, and micrographs, from pre-defined high-quality datasets available on EMPIAR.
Finally, we re-process the raw data using cryoSPARC \cite{cryosparc} through a custom processing pipeline designed to exclude low-quality micrographs and movies, generate annotations for downstream tasks, and verify the resolutions of the reconstructed results.
This workflow has allowed us to compile a large-scale, curated cryo-EM dataset containing over 270,000 raw micrographs and more than 50,000 raw movies from 529 verified single-particle cryo-EM datasets, occupying approximately 25 TB of disk storage in total.
Details of the cryoSPARC processing pipeline are provided in Appendix \ref{sec:pretrain-dataset-detail}.

% 构建高质量的数据集是 \ours 模型训练的关键环节。
% EMPIAR\cite{Iudin2022EMPIARTE}（empiar.org）是一个公开数据库，存储原始了 cryo-EM 图像数据以及 vEM 和 XT 实验的 3D 重建结果。
% 我们从结构生物学的角度出发，定义数据集的质量：高质量的数据集应该能够产出高分辨率的 3D 重建，以便进行后续分析。
% 我们最终选择了分辨率优于 10 \AA 的高质量图像数据集。
% 我们把由完整frames，奇数frames，偶数frames处理得到的single frame micrographs分别记为full、odd、 even single frame micrographs。
% 该数据集包含超过 270,000 张 full single frame micrographs和超过 50,000 组 odd/even single frame micrographs对，涵盖了 529 种不同类型，总存储量超过 25TB。

\noindent\textbf{Data augmentation.} 
Each micrograph in a triplet goes through the same data augmentation process: randomly cropping to between 1/16 and 1/4 of the original size (typically $4096 \times 4096$), resizing to $256 \times 256$, applying random horizontal and vertical flips, and finally normalizing based on the mean and standard deviation computed from the original micrograph.
We randomly crop each micrograph 16 times within a single epoch for fully utilization.

\noindent\textbf{Pre-training details.} 
We explore the performance of \ours using two ViT architectures for the encoder, ViT-B and ViT-L, denoted as \textbf{\ours-B} and \textbf{\ours-L}, respectively.
The decoder of \ours uses 8 Transformer blocks with embedding dimension 512, followed by a three-layer convolution neck and a linear projection layer with an output dimension $16\times16$, which is also the patch size of the input. The mask ratio for the one input micrograph is 0.75 by default.
To fully utilize our large-scale curated dataset, we warm up \ours on the original 270,000 micrographs based on the MAE training scheme for 200 epochs. Then we adopt our novel denoising-reconstruction pre-training for 400 epochs. The warm-up stage takes 6 hours, and the pre-training stage takes 16 hours on a GPU cluster with 64 NVIDIA A800 GPUs, requiring approximately 80 GB of memory for a batch size of 4096.

\subsection{Particle Picking}
% 颗粒挑选是一个在高噪声micrograph中精准识别蛋白质颗粒位置的任务，每张micrograph都可以包含成百上千个蛋白质颗粒，挑选的结果会直接影响蛋白质最终重建的分辨率。
% 我们采用了基于 ViT 的目标检测框架 detectron2 \cite{detectron}，将 \ours 的 encoder 接入框架中全量微调。

Particle picking aims to accurately locate particles in highly noisy micrographs, which is directly related to the resolution of the final reconstructed result.
For adaption, we conduct supervised fine-tuning based on the ViT-based object detection framework Detectron2 \cite{vitdet} on our curated dataset.
The results show that \ours is capable of accurately detecting particles of various shapes and sizes across three challenging datasets.

% 我们利用 cryoSPARC 的标准工作流程生成了一个含有 46 种颗粒超过 80,000 张单帧图像和约 8M 颗粒的高质量标注数据集，用于评估颗粒挑选的效果。我们使用已经解好的蛋白质3D结构作为reference辅助挑选高质量颗粒，详细的工作流程见 Appendix \ref{sec:picking-dataset}。
% 测试数据集包括 EMPIAR ID 为 10081，10350，10407 的三套数据集。

\noindent\textbf{Dataset.} To create the annotated dataset for adaptation, we employ the standard workflow of cryoSPARC to generate a high-quality annotated dataset containing over 80,000 full micrographs and approximately 8 million particles across 46 types of protein.
% The picking of high-quality particles is assisted by using the resolved 3D structures of proteins as references.
Detailed descriptions of the annotation workflow can be found in Appendix \ref{sec:picking-dataset}. The test dataset includes three full micrograph sets with EMPIAR ID 1) 10081: Human HCN1 channel protein 
\cite{10081}, the protein structure has a similar shape to ice, which could lead to false positive picking, 2) 10350: LetB transport protein \cite{10350}, this kind of protein tends to aggregate together, posing challenges in accurate picking in crowded area, and 3) 10407: 70S ribosome \cite{10407}, the micrographs are in the extremely low SNR.

\begin{figure}[t]
  \centering
  % \fbox{\rule[-.5cm]{0cm}{4cm} \rule[-.5cm]{12cm}{0cm}}
  \includegraphics[width=\linewidth]{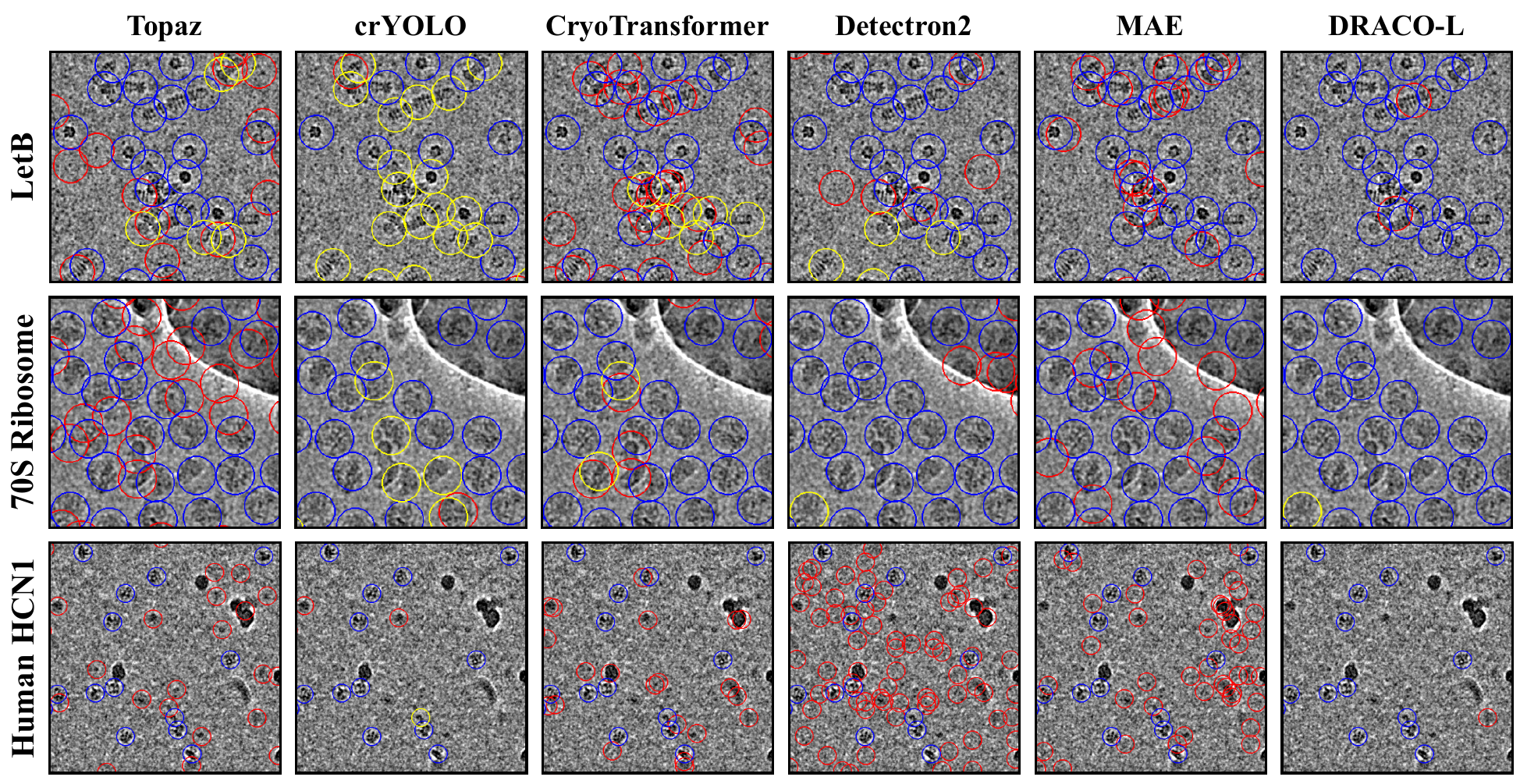}
  \caption{\textbf{Visualization of particle picking results.} We show the picking results of \ours and baselines on the test datasets range from small transport proteins to huge ribosomes. Blue, red, and yellow circles denote true positives, false positives, and false negatives, respectively.}
  \label{fig:particle picking visualize}
\end{figure}

\begin{table}[t]
    \caption{
        \textbf{Particle picking results.} We report the precision, recall, F1 score, and resolution on each test dataset among all baselines. The resolution is obtained from the default cryoSPARC workflow. We compare \ours with existing state-of-the-art methods and consistently achieve the best F1 score and resolution of reconstructed results.
    }
    \label{tab:particle-picking-metrix}
\resizebox{!}{0.44in}{
    \begin{tabular}{c|cccc|cccc|cccc}
        \hline
         & \multicolumn{4}{c|}{\textbf{Human HCN1}} & \multicolumn{4}{c|}{\textbf{70S ribosome}} & \multicolumn{4}{c}{\textbf{LetB}} \\
         \hline
        Method & Precision ($\uparrow$) & Recall ($\uparrow$) & F1 score ($\uparrow$) & Res. ($\downarrow$)&  Precision & Recall  & F1 score  & Res. & Precision & Recall  & F1 score  & Res. \\
        \hline
        Topaz & 0.462 & \textbf{0.956} & 0.623 & 4.20 & 0.362 & \textbf{0.943} & 0.523 & 2.80 & 0.518 & 0.761 & 0.617 & 3.67 \\
        crYOLO & \textbf{0.818} & 0.748 & 0.782 & 4.15 & 0.602 & 0.869 & 0.711 & 2.78 & 0.632 & 0.163 & 0.224 & 4.62 \\
        CryoTransformer & 0.475 & 0.910 & 0.624 & 4.13 & 0.517 & 0.887 & 0.654 & 2.79 & 0.429 & 0.706 & 0.534 & 3.67 \\
        \hline
        Detectron & 0.392 & 0.834 & 0.533 & 4.50 & 0.668 & 0.901 & 0.767 & 2.85 & 0.589 & 0.804 & 0.680 & 3.86 \\
        MAE & 0.703 & 0.649 & 0.675 & 4.32 & 0.712 & 0.876 & 0.786 & 2.84 & 0.591 & \textbf{0.805} & 0.682 & 4.03\\
        \textbf{\ours-B} & 0.768 & 0.799 & 0.793 & 4.03 & 0.732 & 0.905 & 0.810 & 2.61 & 0.637 & 0.779 & 0.701 & 3.55 \\
        \textbf{\ours-L} & 0.830 & 0.802 & \textbf{0.816} & \textbf{3.90} & \textbf{0.803} & 0.846 & \textbf{0.824} & \textbf{2.51} & \textbf{0.678} & 0.780 & \textbf{0.725} & \textbf{3.53} \\
        \hline
        
    \end{tabular}
    }

\end{table}

\noindent\textbf{Baseline and metrics.}
% 我们与现有的颗粒挑选方法如 Topaz\cite{topaz-picking} ，crYOLO\cite{cryolo}，cryoTransformer\cite{cryotransformer}。
% 这些模型均使用了各自的开源模型与默认参数。
% 除了常规的 precision， recall， F1 score 等指标外，重建质量也是 cryo-EM SPA 颗粒挑选的重要指标。
% 我们将挑选出的颗粒接入 cryoSPARC 标准工作流程 \cite{cryosparc}，最终输出重建的density map与分辨率,详见Appendix \ref{sec:picking-recon}。
We compare \ours with existing state-of-the-art learning-based methods for generalized particle picking, including \textbf{Topaz} \cite{topaz-picking}, \textbf{crYOLO} \cite{cryolo}, and \textbf{CryoTransformer} \cite{cryotransformer}.
For ViT-based baselines, including \textbf{MAE}, \textbf{\ours-B} and \textbf{\ours-L} as aforementioned.
We integrate them into Detectron2 framework by loading pre-trained weights of \ours's encoder into Detectron2's encoder.
Additionally, to show the effectiveness of pre-trained encoder, we compare with \textbf{Detectron2} trained from scratch.
More configuration details can be found in Appendix \ref{sec:picking-setting}.
We evaluate baselines in terms of conventional metrics including precision, recall, and F1 score, and the resolution of the resolved 3D structure from the picked particles, which is also a crucial metric.
We process the particles selected by each method with the standard cryoSPARC workflow and finally produce 3D reconstruction density maps and the corresponding resolution, as described in Appendix \ref{sec:picking-recon}. 

\noindent\textbf{Results.}
% 我们的结果如Table\ref{tab:particle-picking-metrix}所示，\ours 在测试数据集上展现出良好的泛化性，重建的分辨率均是最高的。如图\ref{fig:particle picking visualize}所示,Topaz和cryoTransformer都倾向于挑选更多的颗粒，但会有很多错检，假阳性。cryolo挑选的准确率比较高，但是挑选的总数量少，漏检多。
As illustrated in Figure \ref{fig:particle picking visualize}, both Topaz and CryoTransformer tend to pick a larger number of particles, but this often results in many false positives. In contrast, crYOLO achieves higher precision in picking, yet exhibits a higher number of false negatives. Detectron2 trained from scratch and pre-trained MAE both have difficulties in distiguishing signal and noise, leading to a lack of generalizability. In contrast, \ours effectively identifies correct particles, surpassing the performance of all baselines, as demonstrated in Table \ref{tab:particle-picking-metrix}. 
% Also, in Table \ref{tab:particle-picking-metrix}, our results demonstrate that \ours outperforms all other baselines in terms of F1 score and the resolution of reconstructed results. 

\subsection{Micrograph Denoising}

% 抗噪的预训练策略使得 \ours 在去噪任务上表现出天然的适应性。
% 采用预训练阶段相同的网络结构，\ours 可以在不加任何 mask 的任意大小的输入micrograph上直接进行去噪。
Once pre-trained, our model can naturally serve as a generalizable denoiser by directly predicting every patch of input noisy micrograph without any further fine-tuning.

\noindent\textbf{Baseline and metrics.}
% 为评估 \ours 的去噪性能，我们将其与标准 MAE 和 Topaz Denoise \cite{topaz-denoise} 进行了比较。
% 为确保公平比较，我们在同样的奇偶帧数据集上重新训练了 Topaz Denoise for 100 epochs。
% 由于 cryo-EM 图像没有干净的 ground truth，我们采用了与Topaz denoise相同的 SNR 计算方法。
% 具体来说，我们在测试数据集上手动标注了N个信号-背景区域对，对于第i个pair中的信号区域$r_i^s$,$r_i^b$,计算出信号区域的mean $\mu_i^s$,与背景区域的mean $\mu_i^b$和variance $v_i^b$,那么平均SNR的计算为(in dB)：
% 具体来说，对于分成奇偶帧的图像，对奇帧图像做去噪，SNR的计算如下：
To evaluate the effectiveness of \ours on the denoising task, we first compare \ours with the standard MAE trained on the 270,000 micrographs from our large-scale curated dataset with ViT-B as the backbone, denoted as \textbf{MAE}. We further compare with a popular denoising method \textbf{Topaz-Denoise} \cite{topaz-denoise} in cryo-EM. For a fair comparison, we train Topaz-Denoise on our odd-even micrograph dataset for 100 epochs with default settings. Last, we compare with the traditional method \textbf{Low-pass} filtering that has already been integrated into commercial software cryoSPARC \cite{cryosparc}. We utilize the same protocol used in cryoSPARC and set the low-pass cutoff resolution to 20 \AA. As cryo-EM micrographs lack clean ground truth, following Topaz-Denoise, we employ an SNR calculation method that involves 20 manually annotated signal-background region pairs as references. For each $i$-th pair, we calculate the mean and variance for both the signal region $r_i^s$ and the background region $r_i^b$, yielding $\mu_i^s$, $v_i^s$ for the signal, and $\mu_i^b$, $v_i^b$ for the background. The average SNR is then computed in dB as follows:
\begin{equation}
    \text{SNR}=\frac{10}{N}\sum_{i=1}^N\text{log}_{10}\frac{(\mu_i^s-\mu_i^b)^2}{v_i^b}
    \label{equation:SNR}
\end{equation}

\begin{figure}[t]
  \centering
  \includegraphics[width=\linewidth]{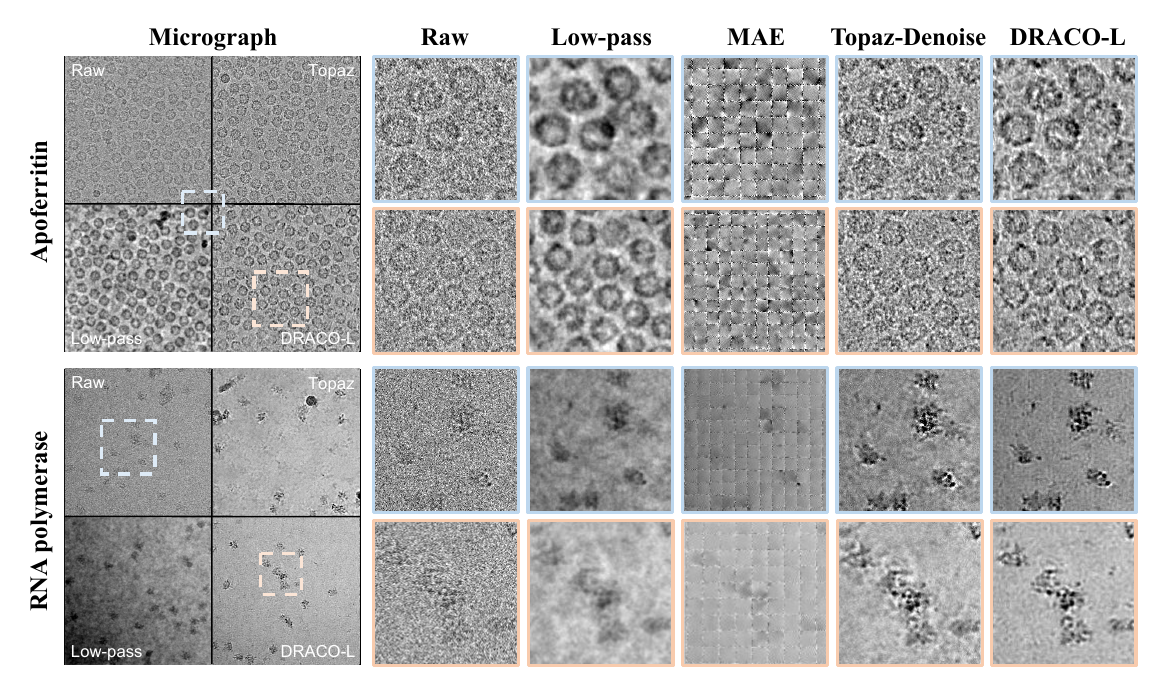}
  \caption{\textbf{Qualitative comparison results of micrograph denoising.} We visualize the denoising results of \ours and state-of-the-art baselines. Our results show the most significant SNR improvement without the loss of the particle structure details. In contrast, Low-pass leads to a severe blur on particles, MAE introduces severe patch-wise artifacts and Topaz only shows either minor SNR improvements or blurred results.}
  \label{fig:denoise}
\end{figure}

\begin{table}[t]
\centering
    \caption{
        \textbf{Quantitative comparison of denoising results.}  We report the SNR calculated with Equation \ref{equation:SNR}. On our test dataset, \ours outperforms all other baselines.}
    \label{tab:denoise metrics}
\resizebox{!}{0.4in}{
    \begin{tabular}{c|cccc} \hline
                    & Human Apoferritin & HA Trimer & Phage MS2 & RNA polymerase \\ \hline
    Raw             &    -10.01    &    -6.52     &   -12.52      &   -4.69      \\
    Low-pass        &    -2.18     &    -0.84     &   -6.71      &   3.09       \\
    Topaz-Denoise   &    -5.67    &    -0.83     &   -6.93      &   8.66       \\ \hline
    MAE             &    -0.31     &    -1.85     &   -8.45      &   1.27       \\
    \textbf{\ours-B}         &    1.92      &    \textbf{3.69}      &   -0.13      &   10.13 \\ 
    \textbf{\ours-L}         &  \textbf{2.01}  &  3.33  &  \textbf{0.23}  & \textbf{12.21} \\
    \hline
    \end{tabular}
    }
\end{table}

\noindent\textbf{Dataset.}
% 我们在训练集外挑选了四套single frame图像数据作为测试数据集以测试去噪效果及泛化性，EMPIAR ID分别为10075，10096，10407，11521，following Topaz denoise, 对于每个数据集，我们都手动对5张图像标注了共20个信号-背景对 for SNR calculation。
To evaluate the denoising capabilities of the baseline models, we select four original micrograph datasets as test datasets, which are \emph{excluded} from the training set.
These sets are Human Apoferritin (EMPIAR-10421) \cite{10421}, HA Trimer (EMPIAR-10096) \cite{10096}, Phage MS2 (EMPIAR-10075) \cite{10075}, and RNA polymerase (EMPIAR-11521) \cite{11521}.
For each dataset, 5 micrographs are selected and 20 signal-background pair regions are labeled in total.
Specifically, these signal and background regions are chosen close together to ensure similar background signals across both regions.

\noindent\textbf{Results.}
% \ours 借助 去噪-重建 的训练策略在去噪任务中表现出卓越的性能，能够有效区分信号与噪声Tab. \ref{tab:denoise metrics}。
% 如图\ref{fig:denoise}所示，标准 MAE 仅能恢复光滑的轮廓，低通滤波虽然平滑了背景噪音，但颗粒的结构信息并没有完全保留住。而 \ours 相较于Topaz Denoise，在保持原有颗粒信号方面表现更好，展示出更好的泛化性。
The quantitative experiments show that \ours achieves significant performance improvements in terms of SNR after denoising compared with state-of-the-art methods, as shown in Table \ref{tab:denoise metrics}.
In Figure \ref{fig:denoise}, the standard MAE can only recover smooth contours of the particle with severe artifacts. Low-pass filtering smooths both signal and background noise, but the background noise is still relatively high and the structure information of particles is corrupted. Topaz sometimes fails to effectively denoise micrographs but over-smooth them instead, which affects the generalizability. In contrast, \ours outperforms all baselines in retaining the original particle signals with the lowest background noises, showing the best generalizability.
We additionally reconstruct 3D density map and corresponding resolution using the denoised particles generated by each method. To ensure that comparisons reflect only the impact of denoising quality, we fix the locations and poses of the picked particles, which were determined using the cryoSPARC workflow. The results are shown in Table \ref{tab:denoise reconstruction metrics}. DRACO consistently achieves the highest resolution in most cases, demonstrating its effectiveness in preserving more high-frequency signals while effectively reducing background noise. 
We demonstrate additional denoising results in Figure~\ref{fig-sup:denoise}.

\begin{table}[t]
\centering
    \caption{
        \textbf{Quantitative comparison of reconstruction using denoised particles.}  We report the resolution obtained from the standard cryoSPARC workflow. On our test dataset, \ours outperforms all other baselines.}
    \label{tab:denoise reconstruction metrics}
\resizebox{!}{0.4in}{
    \begin{tabular}{c|cccc} \hline
                    & Human Apoferritin & HA Trimer & Phage MS2 & RNA polymerase \\ \hline
    Low-pass        &    2.63     &    \textbf{2.06}     &   3.46      &   2.75       \\
    Topaz-Denoise   &    2.34    &    3.06     &   2.52      &   2.93       \\ \hline
    MAE             &    2.77     &    2.15     &   3.78      &   2.81       \\
    \textbf{\ours-B}         &    \textbf{2.05}      &    2.10      &   \textbf{2.51}      &   \textbf{2.56} \\ 
    \hline
    \end{tabular}
    }

\end{table}

\subsection{Micrograph Curation} A modern cryo-EM can capture thousands of micrographs in a day. However, the quality of captured micrographs is unverified. Low-quality micrographs may arise from artifacts such as empty sample, ice crystals, ethane contamination, severe drifting, etc. \cite{miffi}.
Low-quality micrographs can negatively contribute to the final reconstruction results. A reliable automated micrograph curation method can significantly improve the efficiency of the data processing pipeline, resulting in shorter processing time and improved final resolution. We show that \ours can easily adapt to this 2-class classification task by linear probing, and achieving the best performance compared to the state-of-the-art method. Similar to \cite{mae}, we freeze the encoder backbone and train an extra linear classification head.

\noindent\textbf{Dataset.} We manually annotate 1,194 micrographs from original micrograph datasets, assigning a binary label (accept or reject) to each to indicate quality. The dataset comprises 617 high-quality and 577 low-quality micrographs. We divided these micrographs into training and evaluation datasets using an 80\%/20\% split ratio.

\noindent\textbf{Baseline and metric.} We compare \textbf{\ours-B} and \textbf{\ours-L} with linear probing against several baselines: a small \textbf{ResNet18} \cite{resnet}; an existing supervised method \textbf{Miffi} \cite{miffi}, which has been trained on 45,000 annotated data; and the standard \textbf{MAE} with linear probing both pre-trained and adapted on our datasets. The ResNet18 is trained from scratch to show the effectiveness of our pre-training strategy. More configuration details is provided in Appendix \ref{sec:curation-setting}. We report the widely used metrics for classification including precision, recall, F1 score, and accuracy on our test dataset. As Miffi predicts multi-labels on low-quality micrographs, we consider them all as rejections for our metric calculations.

\noindent\textbf{Results.}
% 如 Tab. \ref{fig:curation visualize} 所示，Miffi限于训练数据的不足在测试数据集上缺少泛化性，resnet和MAE均难以将噪声与信号分离以进行判断。\ours 能够更有效地从噪声图像中提取图像的全局信息，从而实现更高的分类准确率，展现出比其他方法更好的泛化性。
As shown in Table \ref{tab:curation-metrics}, Miffi, limited by insufficient training data, lacks generalizability on the test dataset. Both ResNet and MAE face difficulties in effectively separating noise from signal for classification for accurate classification. In contrast, \ours extracts global information from noisy images more effectively, resulting in higher classification accuracy and demonstrating better generalizability compared to other methods.

\begin{table}[t]    
    \caption{\textbf{Quantitative comparison of micrograph curation.} Miffi employs its own general model, while ResNet18 is trained from scratch. \ours reports the best results on all four classification metrics.}
    \label{tab:curation-metrics}
\centering 
\resizebox{!}{0.35in}{
    \begin{tabular}{c|cccc} \hline
        Method & Accuracy & Precision & Recall & F1 score \\ \hline
        Miffi    & 0.836 & 0.899 & 0.845 & 0.871 \\ 
        ResNet18 & 0.938 & 0.923 & 0.960 & 0.940 \\ \hline
        MAE    & 0.904 & 0.927 & 0.892 & 0.909 \\ 
        \textbf{\ours-B}  & 0.963 & \textbf{0.976} & 0.953 & 0.964 \\
        \textbf{\ours-L}  & \textbf{0.983} & \textbf{0.976} & \textbf{0.992} & \textbf{0.984} \\ \hline
    \end{tabular}
}

\end{table}

\subsection{Ablation study}
\begin{table}[t]
    \caption{\textbf{Evalution of mask ratios.} We demonstrate the performance of \ours with different mask ratios. The result shows that at the 0.75 mask ratio, \ours achieves the best performance.}
    \label{tab:curation-metrics-ablation}
\centering
\resizebox{!}{0.35in}{
    \begin{tabular}{c|cccc|cc}
    \hline
                        & \multicolumn{4}{c|}{Micrograph Curation}                          & \multicolumn{2}{c}{Denoising}   \\
    \hline
    Mask Ratio & Accuracy       & Precision      & Recall         & F1 Score       & RNA polymerase  & HA Trimer    \\ \hline
    0.5                 & 0.954          & 0.968          & 0.945          & 0.956          &      10.29      &    2.57       \\
    0.625               & 0.930          & 0.960          & 0.909          & 0.933          & \textbf{10.49}  &    2.80       \\
    0.75                & \textbf{0.963} & 0.976          & \textbf{0.953} & \textbf{0.964} &      10.13      & \textbf{3.69} \\
    0.875               & 0.958          & \textbf{0.984} & 0.939          & 0.961          &      9.59       &    2.28       \\ \hline
    \end{tabular}
}
\end{table}
\begin{table}[t]
    \caption{\textbf{Evalution of loss function.} We demonstrate the performance of \ours with different training objective on the 70S ribosome dataset. The result shows that with both loss, \ours achieves the best performance.}
    \label{tab:loss-function-ablation}
\centering
\resizebox{!}{0.35in}{
    \begin{tabular}{c|cccc|c}
    \hline
                        & \multicolumn{4}{c|}{Particle Picking	}                          & \multicolumn{1}{c}{Denoising}   \\
    \hline
    Training scheme & Precision($\uparrow$)    & Recall($\uparrow$)   & F1 Score($\uparrow$)     & Res.($\downarrow$)   &   SNR($\uparrow$)  \\ \hline
    \textbf{DRACO-B w/o N2N}                 & 0.712          & 0.876          & 0.786         & 2.84         &      -4.94           \\
    \textbf{DRACO-B w/o recon}               & 0.713          & 0.817          & 0.761         & 2.85          & -4.22        \\
    \textbf{DRACO-B}               & \textbf{0.732} & \textbf{0.905}          & \textbf{0.810} & \textbf{2.61} &      \textbf{-2.86}       \\ \hline
    \end{tabular}
}

\end{table}

% 我们同时也比较了\ours 与MAE pretrain在颗粒挑选上的效果，结果在 Tab. \ref{tab:pp ablation study}。我们在EMPIAR-10081上比较了各个预训练策略，可以看到，相比于传统MAE预训练，\ours 提取出的信息在局部特征识别上的表现更加突出。
We compare the performance of networks with different parameter sizes in particle picking and micrograph curation tasks, as shown in Table \ref{tab:particle-picking-metrix} and Table \ref{tab:curation-metrics}.
The results demonstrate that our method can effectively scale up.
Additionally, we evaluate the impact of different mask ratios on denoising and micrograph curation performance, as shown in Table \ref{tab:curation-metrics-ablation}. \ours achieves the highest SNR and curation metric at a 0.75 mask ratio, thus we choose 0.75 as our default mask ratio.
We have also conducted an additional ablation study of loss design. Specifically, we remove either the N2N loss (\textbf{w/o N2N}) or the reconstruction loss (\textbf{w/o recon}) from the training and evaluate the resulting models on particle picking and denoising tasks, as shown in Table \ref{tab:loss-function-ablation}. The result shows that both N2N and reconstruction losses improve performance. Without the N2N loss, \ours struggles to distinguish between signal and noise in micrographs. Without the reconstruction loss, \ours loses its ability to extract general features. We show additional visualization of their denoising results (Figure~\ref{fig-sup:mask-ratio-denoise}) in the Appendix~\ref{sec-supp:add}.

\section{Discussion}\label{sec:conclusion}

\noindent\textbf{Limitations.} As the first attempt to achieve robust feature extraction for cryo-EM via a novel denoising-reconstruction autoencoder, our work presents opportunities for future enhancements. First, our method relies heavily on the performance of motion correction algorithms. This can be improved by designing a more comprehensive denoising task for raw noisy movies. Second, although we have collected what we believe to be the largest curated dataset for cryo-EM, it focuses primarily on mainstream single-particle datasets. To enhance dataset diversity, other types of cryo-EM datasets, such as cryo-electron tomography (cryo-ET)\cite{cryoet} datasets, should also be included. Finally, our current approach only supports various micrograph-level downstream tasks. For particle-level tasks, such as pose estimation, a more fine-grained yet robust feature extraction is required. This can be achieved by developing a particle-level version of \ours{}.
% 1. 基于前置的motion correction算法。
% 2. data scale: 目前只包含了single-particle数据，更多模态的cryo-EM数据可以被引入。
% 3. 对下游任务的支持比较有限。
% particle-level的细节提取比较受限，原因是网络输入的分辨率太低, 可以通过crop particle区域，训练针对于particle结构的DRACO。
% 4. 

\noindent\textbf{Conclusion.} We have introduced \ours, a foundation model designed specifically for cryo-EM image processing, supported by a unique denoising-reconstruction pre-training framework to enable robust feature extraction for cryo-EM micrographs. We have constructed a diverse and high-quality cryo-EM image dataset from the uncurated public database, comprising over 270,000 movies and micrographs. After pre-training, our model's versatility is evidenced by its superior adaptation performance across multiple downstream tasks, including denoising, micrograph curation, and particle picking.
All code, pre-trained model weights, and datasets will be made publicly available for further research and model development.

\section{Acknowledgement}

This work was supported by HPC Platform of ShanghaiTech University. We thank Zhenyang Xu for shaping the paper.

{
    \small
    \bibliography{Chapter/reference}
}
\clearpage
\appendix

\section{Additional Results of \ours}\label{sec-supp:add}

\begin{figure}[htpb]
    \centering
    \includegraphics[width=\linewidth]{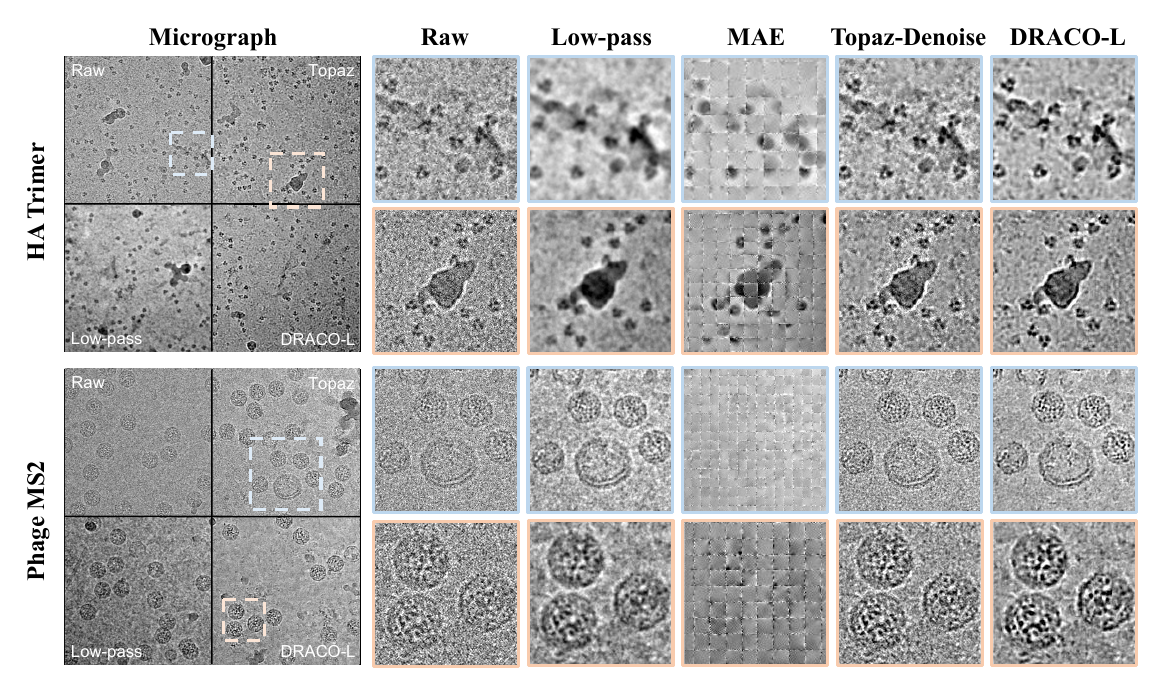}
    \caption{\textbf{Additional denoising results.} We have conducted additional experiments on datasets of membrane proteins and bacteriophages. DRACO achieves the highest visual denoising quality by optimally balancing signal preservation and noise reduction.}
    \label{fig-sup:denoise}
\end{figure}

\begin{figure}[htpb]
    \centering
    \includegraphics[width=\linewidth]{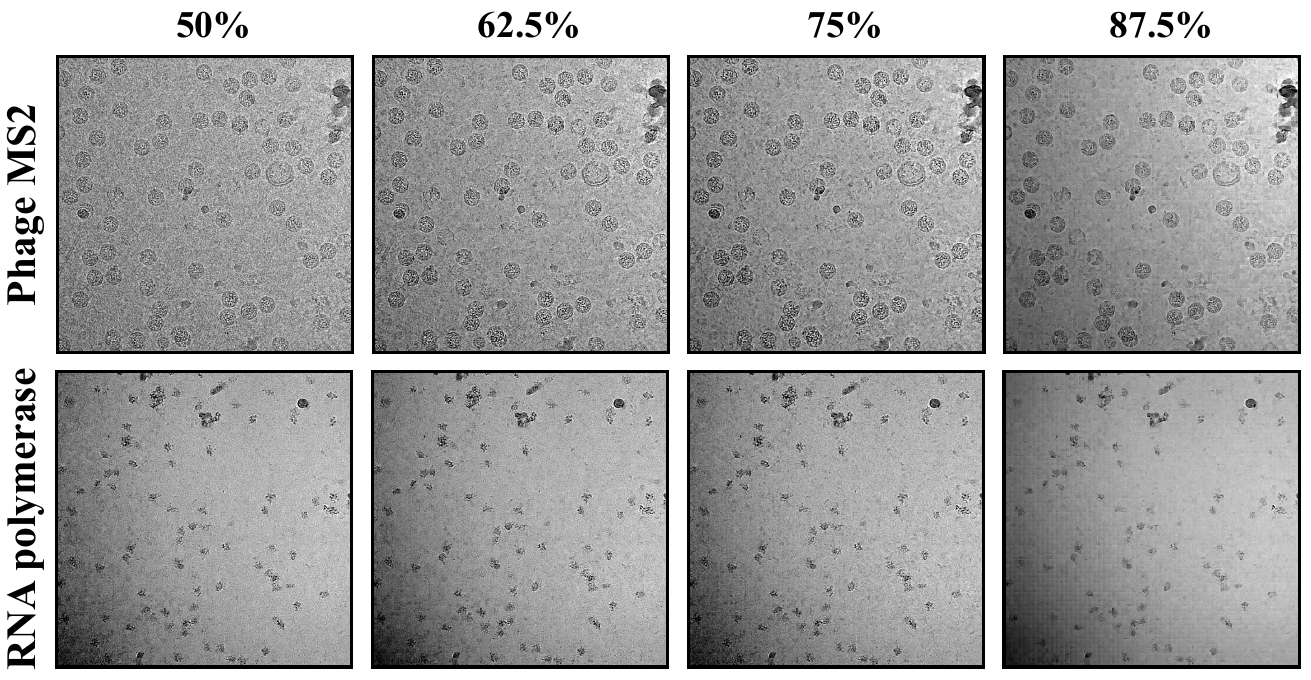}
    \caption{\textbf{Visualization of \ours{}'s results on different mask ratios.} At a 0.75 mask ratio, \ours achieves the best trade-off between signal preservation and background noise removal.}
    \label{fig-sup:mask-ratio-denoise}
\end{figure}

\begin{figure}[htpb]
    \centering
    \includegraphics[width=\linewidth]{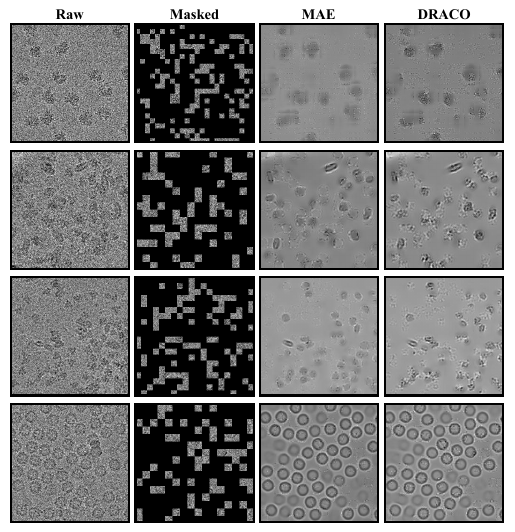}
    \caption{\textbf{Additional results on image reconstruction.} We present the reconstruction results at various image resolutions while maintaining a consistent mask ratio of 0.75. \ours demonstrates enhanced detail preservation on the visible patches compared to MAE.}
    \label{fig-sup:reconstruction}
\end{figure}

We demonstrate that \ours achieves the highest visual denoising quality in terms of both signal preservation and noise removal, as shown in Figure~\ref{fig-sup:denoise}. This figure serves as an extension to Figure \ref{fig:denoise} in our main paper.
Furthermore, we visualize the denoising results from our ablation study on different mask ratios in Figure \ref{fig-sup:mask-ratio-denoise}. We observe that at a 0.75 mask ratio, \ours achieves the best denoising results, as supported by Table~\ref{tab:denoise metrics} in our main paper.
Lastly, we visualize the reconstruction ability at a 0.75 mask ratio in Figure \ref{fig-sup:reconstruction}. \ours demonstrates comparable reconstruction capability to MAE, while significantly better denoising results on visible patches.

\section{Zero-shot Capability on Cryo-ET}\label{sec-supp:cryoet}

\begin{figure}[htpb]
    \centering
    \includegraphics[width=\linewidth]{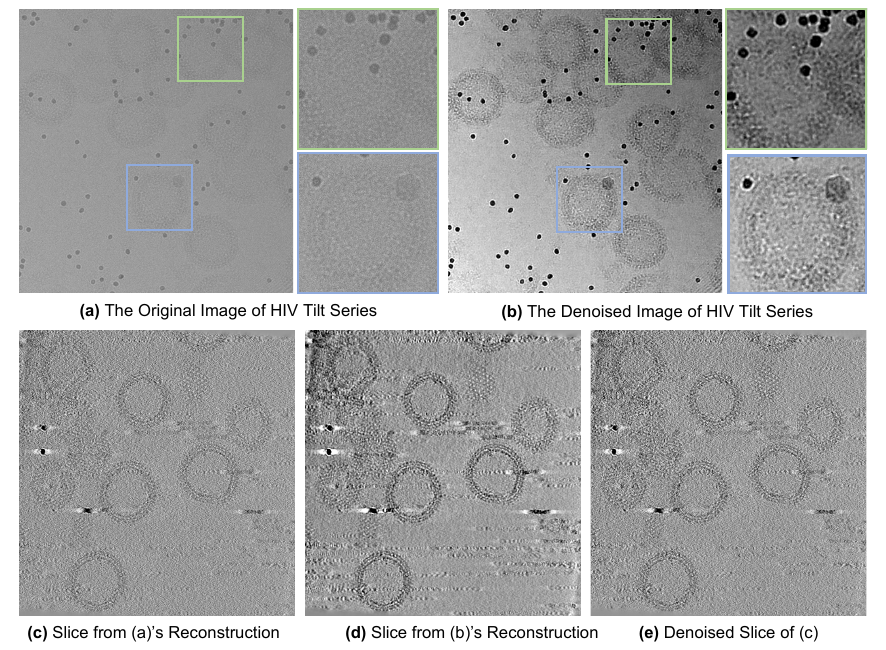}
    \caption{\textbf{Denoising cryo-ET HIV tilt series with \ours.} Figure (a) and (b) show the HIV tilt series beforeand after DRACO’s denoising process. Using \href{https://bio3d.colorado.edu/imod/}{IMOD}, we reconstruct 3D volumes of HIV from both the original and denoised series, showing their slice in Figures (c) and (d). Note the horizontal stripes in these images, which are artifacts due to the missing wedge issue in cryo-ET. Figure (e) shows a denoised slice from Figure (c) by DRACO.}
    \label{fig-sup:cryo-et-denoise}
\end{figure}

Though DRACO has not been pre-trained on cryo-ET datasets, we found that DRACO can be directly applied on cryo-ET tilt series. Here, we demonstrate that DRACO is capable of denoising the unseen HIV tilt series \cite{hiv}, as shown in Figure \ref{fig-sup:cryo-et-denoise}. Specifically, we evaluate DRACO on both the tilt series and the volume slices, showing that DRACO effectively removes background noise and achieves higher contrast in the volume. Furthermore, we assess DRACO's performance in the context of reconstruction. We compare the slices from both tomograms reconstructed using original and denoised tilt series. Additionally, we compare these with the denoised slice from the original tomogram. The result shows that DRACO can improve the contrast of slices before or after the reconstruction.

\section{Workflow of CryoSPARC}\label{sec:cryosparc-pipeline}

% \subsection{Proof of Zero Mean Noise}\label{sec:zero-mean}
% Following the notation in Sec. \ref{sec:preliminary} and leveraging the property of the Poisson distribution where its mean and variance are equal to the parameter of the distribution, we have:
% \begin{equation}
% \begin{aligned}
%     \mathbb{E}[\epsilon] &= \mathbb{E}[\hat{I}_i - I]\\
%     &= \mathbb{E}[\text{Poisson}(I) + \mathcal{G} - I]\\
%     &= \mathbb{E}[\text{Poisson}(I)] + \mathbb{E}[\mathcal{G}] - \mathbb{E}[I]\\
%     &= I + 0 - I\\
%     &= 0.
% \end{aligned}
% \end{equation}
% Thus, under this definition, the noise has a zero mean.

% \subsection{Workflow}\label{sec:cryosparc-pipeline}

\subsection{Pre-training Dataset Details}\label{sec:pretrain-dataset-detail}
% EMPIAR~\cite{Iudin2022EMPIARTE}(empiar.org) 是存储原始cryo-EM图像数据和vEM、XT实验的3D重建的公共档案库,现在包含超过2000个条目，总计超过2PB的数据。
% EMDB (https://emdb-empiar.org) 是由cryo-EM实验衍生的3D重建档案库，它的许多条目补充了EMDB-related的EMPIAR条目的实验信息（样品制备相关，重建流程相关等）。

EMPIAR \cite{empiar} is a public archive for storing raw cryo-EM images and 3D reconstructions from vEM and XT experiments. It currently contains over 2,000 entries, totaling more than 2 PB of data. The EMDB \cite{emdb} is an archive of 3D reconstructions derived from cryo-EM experiments, many of which supplement the experimental information of the EMDB-related EMPIAR entries, such as sample preparation and reconstruction processes.

% 在3D electron microscpy (3DEM)领域中，cryoEM图像进行复杂的图像处理并获得带有分辨率的3D重建，专家能够在分辨率好于3\AA 的3D重建上几乎无错误地搭建蛋白质分子模型从而进行生物学分析， 而在分辨率低于4\AA 的3D重建上避免出错是非常具有挑战性的~\cite{Jamali2024AutomatedMB}， 我们从生物学分析的角度定义数据集的质量，高质量的数据集应该能够产出高分辨的3D重建以进行后续分析，我们选择重建分辨率作为定量指标来筛选高质量图像数据集。

In the field of 3D electron microscopy, cryo-EM micrographs undergo complex image processing to achieve 3D reconstructions with specified resolutions. Experts can model protein molecules accurately on 3D reconstructions with resolutions better than 3 \AA{}. We define the quality of datasets from the perspective of structural biology; hence, high-quality datasets should produce high-resolution 3D reconstructions suitable for detailed analysis.

% 在构建预训练数据集时，我们通过EMPIAR提供的REST API获取了每套数据集的metadata（实验类型，EMDB ID，图像分类等），专门筛选了实验类型是EMDB的EMPIAR 条目，而后获取EMDB条目中的重建信息，优先选择满足重建分辨率优于10\AA 的高质量图像数据集。随后，我们尽可能的收集了符合条件的包含 single frame 的数据集，并对其中包含有 multi frames 的数据集进行了额外处理，分离出奇数帧和偶数帧图像。

When constructing the pre-training dataset, we utilize the REST API provided by EMPIAR to obtain metadata for each dataset, such as experiment type, EMDB ID, and image classification. We specifically filter for EMPIAR entries that are experimentally linked to EMDB, prioritizing those with reconstruction resolutions better than 10 \AA{}. Subsequently, we collect as many datasets as possible that contain single-frame micrographs. For datasets that include multiple frames, we further process them by separating the frames into odd and even micrographs.

% cryo-EM原始数据是记录电子相应数目的multi frames，称为movie。Movies经过运动矫正过后得到的single frame，称为micrograph。我们按照图像分类，尽可能多地下载single frame micrographs，并通过CryoSPARC的Patch Motion Correction将multi frame movie处理为single frame micrograph，对于每一张multi frame movie，我们将完整frames、奇数frames、偶数frames分别处理得到三种single frame micrographs进行N2N学习。

Cryo-EM raw data consists of multi-frame recordings known as movies, which capture the number of electrons. After motion correction \cite{motioncor2}, these movies yield single frames referred to as micrographs. We download as many single-frame micrographs as possible, categorized by image type. Additionally, we process multi-frame movies into single-frame micrographs using cryoSPARC's Patch Motion Correction. For each multi-frame movie, we separately process the complete frames, odd frames, and even frames to generate three types of single-frame micrographs for \ours learning.

\subsection{Particle Picking Dataset Details.}\label{sec:picking-dataset}
% 我们筛选并生成了46套约80k张single frame约8M颗粒标注的数据集，具体过程基于Cryo-EM SPA重建管线软件Cryosparc[xxx]。首先，每个EMPIAR公开数据集都有已经解好的3D density map，which is available on EMDB,使用"create template"步骤将其以50个pose投影，可以得到各个角度高质量的template用于template picking，再经过连续两次2D classification去除可能存在的假阳性颗粒，最终，用这些颗粒重建出分辨率与已经解好的结构相差不超过20\%的结果，就能收集到高质量的颗粒标注数据集。

We filter and generate a dataset comprising approximately 80,000 single-frame micrographs annotated with about 8 million particles, following a process based on the cryo-EM single particle analysis reconstruction pipeline software, cryoSPARC \cite{cryosparc}. Each EMPIAR public dataset \cite{empiar} comes with a solved 3D density map, available on EMDB \cite{emdb}. Using the ``create template'' step in cryoSPARC, we project this map into 50 diverse poses to generate high-quality templates for template picking. Subsequent rounds of the ``2D classification'' step are employed to eliminate potential false positives. Finally, using these particles, we reconstruct results whose resolution did not differ by more than 20\% from the reported resolution. This method allowed us to collect a high-quality annotated particle dataset.

\subsection{3D Reconstruction Pipeline}\label{sec:picking-recon}
% 重建的流程同样基于Cryo-EM SPA重建管线软件Cryosparc[xxx].在获得挑选出的颗粒后，我们先经过一次2D classification去除假阳性颗粒，做一次默认参数的ab initio reconstruction重建初始3D model，基于初始模型，再做一次homogeneous refinement以重建高分辨率结果。分辨率是由Fourier shell correlation(FSC)曲线决定，具体做法是，将particle分为随机的两个half，分别做homogeneous重建，重建完的3D density map在频域上的每个fourier shell上做cross correlation，最终得出每个fourier shell对应的correlation值，以标准的0.143位threshold得到最终的分辨率。

The reconstruction process is also based on cryoSPARC. After picking the particles, the standard reconstruction workflow consists of 2D classification, ab initio reconstruction and homogeneous refinement. 2D classification aims to remove any false positives in picked particles. Ab initio reconstruction can create an initial 3D model from a certain set of particles. Based on this initial model, homogeneous refinement can achieve a high-resolution result. The final resolution is determined by the Fourier shell correlation (FSC) curve. The specific method involves dividing the particles into two random halves, each undergoing homogeneous reconstruction. After reconstruction, we perform a cross-correlation on each Fourier shell in the frequency domain of two reconstructed 3D density maps. The final resolution is determined using the standard threshold of 0.143 on the FSC curve.

\section{Downstream tasks settings}

\subsection{Particle Picking Settings}\label{sec:picking-setting}

\paragraph{Particle picking baselines.}
We use the Topaz \cite{topaz-picking} general model with its ``resnet16u64'' backbone for our baseline, picking particles that score higher than 0.0 as the final results.
We use the crYOLO \cite{cryolo} general model for our baseline from its \href{https://cryolo.readthedocs.io}{official website}.
Similarly, we use CryoTransformer's \cite{cryotransformer} open-source general model on ~\href{https://github.com/jianlin-cheng/CryoTransformer}{Github}, choosing particles with scores ranging from the 25th to the 100th percentile.

\paragraph{\href{https://github.com/facebookresearch/detectron2}{Detectron2} configurations.}
For particle picking, we employ the Faster R-CNN framework within Detectron2, to fit the particle picking task with our curated dataset.
The configurations for both ViT-B and ViT-L include the standard feature pyramid and window attention \cite{swin}.
We also set the non-maximum suppression threshold of the region proposal network to 0.6 and adjust the pooling size of the box pooler in the region of interest network to 14.
Given that the particles are mostly square, the aspect ratio of the anchors is fixed to 1.0.
All other settings remain at their default values.
The data augmentation goes the same process as described in the pre-training stage.
We fine-tune Detectron2-based particle picking model on 64 NVIDIA A800 GPUs for 100 epochs with a batch size of 256, requiring approximately 9 hours and consuming around 100GB of memory.
After fine-tuning, we process the test dataset and pick particles with scores higher than 0.1 as the final results.

\subsection{Micrograph Curation Settings}\label{sec:curation-setting}
For micrograph curation, each ViT-based model undergoes a linear probing phase, while a ResNet18 \cite{resnet} is trained from scratch. We employ the miffi\_v1 \cite{miffi} general model of Miffi to inference on test datasets.
All the model trains 50 epochs with a batch size of 128 on a single NVIDIA RTX 3090 GPU, taking about 10 minutes and utilizing around 8GB of memory.

\end{document}